\documentclass[12pt,english]{scrartcl}
\usepackage[T1]{fontenc}
\usepackage[latin9]{inputenc}
\usepackage{geometry}
\usepackage{color}
\usepackage{babel}
\usepackage{array}
\usepackage{multirow}
\usepackage{amsmath}
\usepackage{amsthm}
\usepackage{setspace}
\usepackage[authoryear]{natbib}
\usepackage[unicode=true,
 bookmarks=true,bookmarksnumbered=true,bookmarksopen=true,bookmarksopenlevel=10,
 breaklinks=true,pdfborder={0 0 0},pdfborderstyle={},backref=section,colorlinks=true]
 {hyperref}
\hypersetup{pdftitle={A Generalization Method of Partitioned Activation Function for Complex Number},
 pdfauthor={Lee, HyeonSeok},
 pdfsubject={Complex Activation Function},
 pdfkeywords={Complex Activation Function, Holomorphic, Phase-preserving,  real-complex interaction}}

\makeatletter

\providecommand{\tabularnewline}{\\}

\DeclareMathOperator{\sinc}{sinc}

\usepackage{breqn}
\gdef\wrap@breqn@environ#1#2{
    \expandafter\let\csname breqn@oldbegin@#1\expandafter\endcsname\csname #1\endcsname
    \expandafter\let\csname breqn@oldend@#1\expandafter\endcsname\csname end#1\endcsname
    \expandafter\gdef\csname breqn@begin@#1\endcsname{%
        \expandafter\let\csname #1\expandafter\endcsname\csname breqn@oldbegin@#1\endcsname%
        \begin{#2}%
    }
    \expandafter\gdef\csname breqn@end@#1\endcsname{%
        \expandafter\let\csname end#1\expandafter\endcsname\csname breqn@oldend@#1\endcsname%
        \end{#2}%
        \expandafter\let\csname #1\expandafter\endcsname\csname breqn@begin@#1\endcsname%
        \expandafter\let\csname end#1\expandafter\endcsname\csname breqn@end@#1\endcsname%
    }
    \expandafter\let\csname #1\expandafter\endcsname\csname breqn@begin@#1\endcsname
    \expandafter\let\csname end#1\expandafter\endcsname\csname breqn@end@#1\endcsname
}
\wrap@breqn@environ{equation}{dmath}
\wrap@breqn@environ{equation*}{dmath*}

\makeatother

\begin{document}
\global\long\def\sinc{\mathrm{sinc}}

\global\long\def\Si{\mathrm{Si}}

\global\long\def\Heaviside{\mathrm{H}}

\global\long\def\erf{\mathrm{erf}}

\global\long\def\logistic{\mathrm{L}}

\title{A Generalization Method of Partitioned Activation Function for Complex
Number }

\author{HyeonSeok Lee \thanks{1st author: Yonsei Univ, Seoul, Korea}, Hyo
Seon Park \thanks{Corresponding Author: Yonsei Univ, Seoul, Korea, hspark@yonsei.ac.kr}}
\maketitle
\begin{abstract}
A method to convert real number \emph{partitioned} activation function
into complex number one is provided. The method has 4em variations;
1 has potential to get holomorphic activation, 2 has potential to
conserve complex angle, and the last 1 guarantees interaction between
real and imaginary parts. The method has been applied to LReLU and
SELU as examples. 

The complex number activation function is an building block of complex
number ANN, which has potential to properly deal with complex number
problems. But the complex activation is not well established yet.
Therefore, we propose a way to extend the partitioned real activation
to complex number. 
\end{abstract}

\section{Motivation}

Complex Number Artificial Neural Net (CANN) is a natural consequence
of using ANN for complex number, while complex number appear in many
problems. The complex number problems include (Fourier, Laplace, Z,
etc) transform based methods \citep{haberman2013applied} , steady-state
problems \citep{banerjee1994theboundary} , mapping 2D problem to
complex plane \citep{greenberg1998advanced} , and electromagnetic
signals such as MRI image \citep{virtue2017betterthan} . Due to such
necessity, CANN appears from 1990's even long before the days of deep
learning, as discussed in \citep{reichert2013neuronal} . As the result
of various efforts, building blocks of CANN has been developed; Convolution,
Fully-connection, Back-propagation, Batch-normalization, Initialization,
Pooling, and Activation. Among them, Complex Convolution and Fully-connection
are essentially multiplication and addition, which are clearly established
for complex number in mathematics \citep{trabelsi2017deepcomplex}
. Back-propagation for complex number is developed for general complex
number function \citep{georgiou1992complex,nitta1997anextension},
and for holomorphic function \citep{lacorte2014newtons} . Complex
Batch-normalization is also developed following the idea of real Batch-normalization
\citep{trabelsi2017deepcomplex} . The complex number Initialization
is suggested to be done in polar-form \citep{trabelsi2017deepcomplex}
, not Cartesian. Max Pooling for complex number is another necessity,
but the authors cannot see any research. Complex Activation is in
active development \citep{arjovsky2015unitary,guberman2016oncomplex,virtue2017betterthan,georgiou1992complex};
see \citep{lacorte2014newtons} for brief review about various types
of complex activation functions. 

For complex activation, holomorphic (a.k.a. analytic) function has
been an important topic \citep{hitzer2013nonconstant,jalab2011newactivation,vitagliano2003generalized,kim2001complex,mandic2009complex,tripathi2010highdimensional}
, since it makes the back-propagation simpler thus training becomes
faster \citep{amin2011wirtinger,lacorte2014newtons2} . The importance
of Complex Phase Angle of CANN has been suggested, including similarity
to biological neuron \citep{reichert2013neuronal}. The importance
leads to the idea of \emph{phase-preserving} complex activation function
\citep{georgiou1992complex,virtue2017betterthan}. In contrast, \citep{hirose2013complexvalued,kim2003approximation}
suggested that \emph{phase-preservation} makes training more difficult. 

We focus on the complex activation, and propose a methods to generalize
\emph{partitioned} activation such as LReLU and SELU for complex number.
There are 4 variations in the method to accommodate various cases;
1 of them is potentially holomorphic and alter phase, 1 is not holomorphic
and alter phase while guarantees interaction between real and imaginary
parts, and 2 are not holomorphic and potentially keep complex phase
angle. 

This article is structured as follows. Sec. \ref{sec:ComplexActivation Review}
is a brief review about existing Complex Activation functions. Then
Sec. \ref{sec:Generalizing Activation} presents our generalization
method along with examples of LReLU \citep{maas2013rectifier} and
SELU \citep{klambauer2017selfnormalizing} . Finally, Sec. \ref{sec:Conclusion}
summarize the article.

\section{Review of Current Complex Activation Functions \label{sec:ComplexActivation Review}}

Certain complex-argument activation functions are derived from real-argument
counterparts, using 3 ways depending on the characteristic of each
function; No change, Modification, and generalization. We will briefly
discuss them with particular weight on the generalization, since our
approach is to generalize. 

\subsection{No change }

Certain no-partitioned activation functions can be used for complex-argument
without any change. Such functions include logistic, $\tanh$, $\tan^{-1}$,
etc. These are already defined for complex-argument, and we can just
use them. But they suffer from poles near the origin \citep{lacorte2014newtons}
. 

\subsection{Modification}

Partitioned activation functions are not straightforward for complex-argument,
due to the partition points on the real-axis. So approach of ``just
take the idea of real activation, and create a corresponding complex
activation'' is tried. We call it \emph{modification}. One example
is modReLU \citep{arjovsky2015unitary} as 

\begin{equation}
\mathrm{modReLU}(z)=\mathrm{ReLU}\left(\lvert z\rvert+b\right)\mathrm{e}^{\mathrm{i}\,\theta}=\begin{cases}
\left(\lvert z\rvert+b\right)\dfrac{z}{\lvert z\rvert} & \lvert z\rvert+b\geq0\\
0 & \lvert z\rvert+b<0
\end{cases}
\end{equation}

where $z\in C$, $\theta$ is the phase angle of $z$, and $b\in R$
is a trainable parameter. The modReLU takes the idea of inactive region
from ReLU, and forms region around origin. Unfortunately, the modReLU
is not holomorphic, while keeps the complex phase. 

\subsection{Generalization}

We call a complex activation \emph{Generalized} , if the values on
real-axis coincide with the real-argument counter-part. The easiest
generalization of partitioned activation function would be to separately
apply activation function to real and imaginary parts \citep{nitta1997anextension,faijulamin2009singlelayered}
. One example is  Separate Complex ReLU (SCReLU) as 

\begin{equation}
\mathrm{SCReLU}(z)=\mathrm{ReLU}\left(x\right)+\mathrm{i}\,\mathrm{ReLU}\left(y\right)
\end{equation}

where $z=x+\mathrm{i}\,y$ . Another simple generalization of ReLU
is to activate only when both real and imaginary parts are positive,
which is called zReLU \citep{guberman2016oncomplex}

\begin{equation}
\mathrm{zReLU}(z)=\begin{cases}
z & x>0,\,y>0\\
0 & otherwise
\end{cases}
\end{equation}

Both SCReLU and ReLU are holomorphic, and are essentially not complex
but 2 separate real activations \citep{lacorte2014newtons} . Another
generalization of ReLU is Complex Cardioid (CC) \citep{virtue2017betterthan}

\begin{equation}
\mathrm{CC}(z)=\frac{1}{2}\left\{ 1+\cos\theta\right\} \,z\label{eq:CC, Complex Cardioid}
\end{equation}

which keeps the complex angle, but is not holomorphic. Our approach
is inspired by the CC. 

\section{New Generalization Method \label{sec:Generalizing Activation}}

Consider a \emph{partitioned} activation function for \emph{real}
number, which has the typical form 

\begin{equation}
f(x)=\begin{cases}
f_{0}(x) & x<0\\
f_{1}(x) & x\geq0
\end{cases}\label{eq:Activation, Partitioned}
\end{equation}

where $f_{0}$ and $f_{1}$ are local functions for positive and negative
regions. Activation functions like LReLU  and SELU  are examples
of the above case. Now, replace partitions with Heaviside unit-step
function $\Heaviside(x)$, and we can rewrite $f\left(x\right)$ as 

\begin{equation}
f(x)=f_{0}(x)\,\Heaviside(-x)+f_{1}(x)\,\Heaviside(x)\label{eq:Activation, Step}
\end{equation}

Next, select a complex-argument function, whose \emph{real} axis values
coincide with the $\Heaviside(x)$ and $\Heaviside(-x)$ . We pick 

\begin{subequations} 

\begin{equation}
\frac{1}{2}\left(1+\mathrm{e}^{\mathrm{i}\,\left(2n+1\right)\theta}\right)\rightarrow\Heaviside(x)
\end{equation}

\begin{equation}
\frac{1}{2}\left(1-e^{\mathrm{i}\,\left(2n+1\right)\theta}\right)\rightarrow\Heaviside(-x)
\end{equation}

\end{subequations} 

where $\theta$ is the phase angle of complex number $z=x+\mathrm{i}\,y$,
and $n\in Z$ (is an integer) is a parameter. Then, we can get a generalized
complex-argument function 

\begin{equation}
f(z)=\frac{1}{2}\left(1-\mathrm{e}^{\mathrm{i}\,\left(2n+1\right)\theta}\right)\,f_{0}\left(z\right)+\frac{1}{2}\left(1+\mathrm{e}^{\mathrm{i}\,\left(2n+1\right)\theta}\right)\,f_{1}\left(z\right)\left(z\right)\label{eq:Generalized Activation 2}
\end{equation}

This approach can be easily extended to more complicated cases, like
S-shaped ReLU (SReLU)  which has 3 partitions. The typical form 

\begin{equation}
f(x)=\begin{cases}
f_{0}(x) & x<x_{0}\\
f_{1}(x) & x_{0}<x\leq x_{1}\\
 & \vdots\\
f_{n-1}(x) & x_{n-2}<x\leq x_{n-1}\\
f_{n}(x) & x_{n-1}<x
\end{cases}\label{eq:Activation, Partitioned 3}
\end{equation}

is rewritten as 

\begin{equation}
f(x)=f_{0}(x)\,\Heaviside(x-x_{0})+f_{1}(x)\,\left\{ \Heaviside(x-x_{0})-\Heaviside(x-x_{1})\right\} +\cdots+f_{n-1}(x)\,\left\{ \Heaviside(x-x_{n-2})-\Heaviside(x-x_{n-1})\right\} +f_{n}(x)\,\Heaviside(-x+x_{n-1})\label{eq:Activation many, Step}
\end{equation}

, then generalized to 

\begin{equation}
f(z)=\frac{1}{2}\left(1-\mathrm{e}^{\mathrm{i}\,\left(2n+1\right)\theta_{0}}\right)\,f_{0}\left(z\right)+\frac{1}{2}\left(\mathrm{e}^{\mathrm{i}\,\left(2n+1\right)\theta_{1}}-\mathrm{e}^{\mathrm{i}\,\left(2n+1\right)\theta_{0}}\right)\,f_{1}\left(z\right)+\cdots+\frac{1}{2}\left(\mathrm{e}^{\mathrm{i}\,\left(2n+1\right)\theta_{n-1}}-\mathrm{e}^{\mathrm{i}\,\left(2n+1\right)\theta_{n-2}}\right)\,f_{n-1}\left(z\right)+\frac{1}{2}\left(1+\mathrm{e}^{\mathrm{i}\,\left(2n+1\right)\theta_{n-1}}\right)f_{n}\left(z\right)\label{eq:Generalized Activation many}
\end{equation}

where $\theta_{p}$ denotes the phase angle of complex number $z_{p}=z-x_{p}=\left(x-x_{p}\right)+\mathrm{i}\,y$,
and $x_{p}$ is the location of partition boundary p. 

In case real valued scale is preferred (eg: to keep the complex angle),
we made simple modification for making replacement of $\Heaviside(x)$
real valued. The modifications are 

\begin{subequations} 

\begin{equation}
\frac{1}{2}\left[1+\cos\left\{ \left(2n+1\right)\theta\right\} \right]\rightarrow\Heaviside(x)
\end{equation}

and 

\begin{equation}
\frac{1}{2}\lvert1+e^{\mathrm{i}\,\left(2n+1\right)\theta}\rvert\rightarrow\Heaviside(x)
\end{equation}

\end{subequations} 

and generalization using them are 

\begin{subequations} 

\begin{equation}
f(z)=\frac{1}{2}\left[1-\cos\left\{ \left(2n+1\right)\theta\right\} \right]\,f_{0}\left(z\right)+\frac{1}{2}\left[1+\cos\left\{ \left(2n+1\right)\theta\right\} \right]\,f_{1}\left(z\right)\label{eq:Generalized Activation 2 cos}
\end{equation}

and 

\begin{equation}
f(z)=\frac{1}{2}\lvert1-e^{\mathrm{i}\,\left(2n+1\right)\theta}\rvert\,f_{0}\left(z\right)+\frac{1}{2}\lvert1+e^{\mathrm{i}\,\left(2n+1\right)\theta}\rvert\,f_{1}\left(z\right)\left(z\right)\label{eq:Generalized Activation 2 abs}
\end{equation}

\end{subequations} 

respectively. 

Another approach is to \emph{approximate} $\Heaviside(x-x_{p})$ on
real axis with a complex-argument function. A few well known such
functions are based on $\tanh$ (hyperbolic tangent) , $\tan^{-1}$
(arc tangent), $\Si$ (sine integral), and $\erf$ (error function)
functions. Among them, we pick $\erf$ based Sigmoid function as below,
since $\tanh$ and $\tan^{-1}$ has poles on imaginary axis, and $\Si$
is oscillatory on real axis. 

\begin{equation}
\mathrm{S}(z)=\frac{1}{2}\left\{ 1+\erf\text{\ensuremath{\left(\dfrac{z}{\sqrt{2}\:\sigma}\right)}}\right\} 
\end{equation}

where $\sigma>0$ is a just a parameter such that smaller $\sigma$
more closely approximates $\Heaviside(x)$ on the real axis. Then
\emph{approximately generalized} functions are 

\begin{equation}
\widetilde{f}(z)=\mathrm{S}(-z)\,f_{0}\left(z\right)+\mathrm{S}(z)\,f_{1}\left(z\right)\label{eq:Approx Generalized Activation 2}
\end{equation}

\begin{equation}
\widetilde{f}(z)=\mathrm{S}(-z_{0})\,f_{0}\left(z\right)+\left(\mathrm{S}(z_{1})-\mathrm{S}(z_{0})\right)\,f_{1}\left(z\right)+\cdots+\left(\mathrm{S}(z_{n-1})-\mathrm{S}(z_{n-2})\right)\,f_{n-1}\left(z\right)+\mathrm{S}(z_{n})\,f_{n}\left(z\right)\label{eq:Approx Generalized Activation many}
\end{equation}

respectively for \eqref{eq:Approx Generalized Activation 2} and \eqref{eq:Approx Generalized Activation many}
cases. An important advantage of this approximate generalization is
that the $\widetilde{f}(z)$ is \emph{holomorphic} , if all the $f_{p}\left(z\right)$
are \emph{holomorphic}. The reason is simple; a) logistic function
is holomorphic, b) product and sum of holomorphic functions are also
holomorphic. Then, each terms in \eqref{eq:Approx Generalized Activation 2}
and \eqref{eq:Approx Generalized Activation many} are holomorphic
which are products of holomorphic, and $\widetilde{f}(z)$ is holomorphic
which is sum of holomorphic. With using normalization (eg: Batch-Renormalization
\citep{ioffe2017batchrenormalization}) is suggested to prevent exploding
feature values toward $\pm\,\mathrm{i}\,\infty$. 

\subsection{Example 1: Leaky Rectified Linear Unit (LReLU)}

The LReLU \citep{maas2013rectifier} is the one of the most popular
activation function (The ReLU is just a LReLU with $\alpha=0$). The
real-argument LReLU is 

\begin{equation}
\mathrm{LReLU}(x)=\begin{cases}
\alpha x & x<0\\
x & x\geq0
\end{cases}\label{eq:LReLU}
\end{equation}

and is generalized using \eqref{eq:Generalized Activation 2} with
n=0 to ``Complex LReLU'' (CLReLU) as

\begin{equation}
\mathrm{CLReLU}(z)=\frac{1}{2}\left\{ \left(1+\alpha\right)+\left(1-\alpha\right)\,\mathrm{e}^{\mathrm{i}\,\theta}\right\} \,z\label{eq:CLReLU}
\end{equation}

which alters both phase and magnitude of input $z$ and has cross-influence
between real and imaginary components. To keep the complex angle of
argument, we use \eqref{eq:Generalized Activation 2 cos} and \eqref{eq:Generalized Activation 2 abs}
with n=0 to get 

\begin{subequations} 

\begin{equation}
\mathrm{cLReLU}(z)=\frac{1}{2}\left\{ \left(1+\alpha\right)+\left(1-\alpha\right)\,\cos\theta\right\} \,z\label{eq:cLReLU}
\end{equation}

and 

\begin{equation}
\mathrm{aLReLU}(z)=\frac{1}{2}\left(\alpha\lvert1-e^{\mathrm{i}\,\left(2n+1\right)\theta}\rvert+\lvert1+e^{\mathrm{i}\,\left(2n+1\right)\theta}\rvert\right)\,z\label{eq:aLReLU}
\end{equation}

\end{subequations} 

which we call ``cos LReLU'' (cLReLU) and ``abs LReLU'' (aLReLU),
respectively. Meanwhile, a ``Holomorphic LReLU'' (HLReLU) can be
derived using \eqref{eq:Approx Generalized Activation 2} as 

\begin{equation}
\mathrm{HLReLU}(x)=\frac{1}{2}\left\{ \left(1+\alpha\right)+\left(1-\alpha\right)\,\erf\text{\ensuremath{\left(\dfrac{z}{\sqrt{2}\:\sigma}\right)}}\right\} \,z\label{eq:HLReLU}
\end{equation}

which is holomorphic, since both the $f_{0}\left(z\right)=\alpha z$
and $f_{1}\left(z\right)=z$ are polynomials which are all holomorphic.
Please note that the HLReLU alters phase angle of argument, since
the argument z is multiplied with a complex number. 

\subsection{Example 2: Scaled Exponential Linear Unit (SELU)}

The SELU \citep{klambauer2017selfnormalizing} was recently developed,
and rapidly becomes popular due to its self-normalizing feature (The
ELU is just a SELU with $\lambda=1$ ). The real-argument SELU is 

\begin{equation}
\mathrm{SELU}(x)=\lambda\begin{cases}
\alpha(\mathrm{e}^{x}-1) & x<0\\
x & x\geq0
\end{cases}\label{eq:SELU}
\end{equation}

where $\lambda>1$and $\alpha$ are parameters with optimum values
$\lambda=1.0507$ and $\alpha=1.67326$. The SELU is generalized using
\eqref{eq:Generalized Activation 2}, \eqref{eq:Generalized Activation 2 cos},
and \eqref{eq:Generalized Activation 2 abs} with n=0 to ``Complex
SELU'' (CSELU), ``cos SELU'' (cSELU) and, ``abs SELU'' (aSELU),
respectively as 

\begin{subequations} 

\begin{equation}
\mathrm{CSELU}(z)=\frac{\lambda}{2}\left\{ \left(1-\mathrm{e}^{\mathrm{i}\,\theta}\right)\alpha(\mathrm{e}^{z}-1)+\left(1+\mathrm{e}^{\mathrm{i}\,\theta}\right)\,z\right\} \label{eq:CSELU}
\end{equation}

\begin{equation}
\mathrm{cSELU}(z)=\frac{\lambda}{2}\left\{ \left(1-\cos\theta\right)\alpha(\mathrm{e}^{z}-1)+\left(1+\cos\theta\right)\,z\right\} \label{eq:cSELU}
\end{equation}

\begin{equation}
\mathrm{aSELU}(z)=\frac{\lambda}{2}\left\{ \lvert1-\mathrm{e}^{\mathrm{i}\,\theta}\rvert\alpha(\mathrm{e}^{z}-1)+\lvert1+\mathrm{e}^{\mathrm{i}\,\theta}\rvert\,z\right\} \label{eq:aSELU}
\end{equation}

\end{subequations} 

Complex phase is altered with all 3 CSELU, cSELU, and aSELU, unlike
cLReLU \eqref{eq:cLReLU} and aLReLU \eqref{eq:aLReLU}. Instead,
all has interaction between real and imaginary components of argument. 

Meanwhile, a Holomorphic ``SELU'' (HSELU) can be derived using \eqref{eq:Approx Generalized Activation 2}
as 

\begin{equation}
\mathrm{HSELU}(z)=\lambda\left\{ \mathrm{S}(-z)\,\alpha(\mathrm{e}^{z}-1)+\mathrm{S}(z)\,z\right\} \label{eq:HSELU}
\end{equation}

which is also holomorphic, since both the scaled shifted exponential
$f_{0}\left(z\right)=\lambda\alpha(\mathrm{e}^{z}-1)$ and polynomial
$f_{1}\left(z\right)=\lambda z$ are holomorphic. 

\section{Concluding Remark \label{sec:Conclusion} }

A generalization of \emph{partitioned} real number activation function
to complex number has been proposed. The generalization process has
2 steps; a) Replace partition on activation with $\Heaviside(x)$,
b) Generalize $\Heaviside(x)$ with a complex number function. The
generalization has 4 variations, and 1 of them is potentially holomorphic.
The generalization scheme has been demonstrated using 2 popular partitioned
activations; LReLU and SELU. The properties of generalized complex
activations are summarized on table \eqref{tab:ComplexActivationSummary}
. 

\begin{table}
\noindent %
\begin{tabular}{|>{\centering}p{2.5cm}|>{\centering}p{4cm}|c|>{\centering}p{3cm}|>{\centering}p{2cm}|}
\hline 
Original Real Activation & Generalized Complex Activation & Holomorphic & Real-Complex Interaction & Phase Preverving\tabularnewline
\hline 
\hline 
\multirow{4}{2.5cm}{LReLU \eqref{eq:LReLU}} & CLReLU \eqref{eq:CLReLU} & X & O & X\tabularnewline
\cline{2-5} 
 & cLReLU \eqref{eq:cLReLU} & X & X & O\tabularnewline
\cline{2-5} 
 & aLReLU \eqref{eq:aLReLU} & X & X & O\tabularnewline
\cline{2-5} 
 & HLReLU \eqref{eq:HLReLU} & O & O & X\tabularnewline
\hline 
\multirow{4}{2.5cm}{SELU \eqref{eq:SELU}} & CSELU \eqref{eq:CSELU} & X & O & X\tabularnewline
\cline{2-5} 
 & cSELU \eqref{eq:cSELU} & X & O & X\tabularnewline
\cline{2-5} 
 & aSELU \eqref{eq:aSELU} & X & O & X\tabularnewline
\cline{2-5} 
 & HSELU \eqref{eq:HSELU} & O & O & X\tabularnewline
\hline 
\end{tabular}

\caption{\label{tab:ComplexActivationSummary} Properties of the Generalized
Complex Activation}

\end{table}

Furthermore, the method can be used for various partitioned real activation
to make them into complex activation. He hope that this humble research
adds another building block for complex ANN, which is important for
complex number problems. 

\bibliographystyle{plainnat}
\bibliography{ComplexActivation}
 
\end{document}